\documentclass[sigconf,10pt,nonacm]{acmart}

\usepackage{amsmath}        
\usepackage{amsfonts}       
\usepackage{amsthm}         
\usepackage{mathtools}      

\usepackage{algorithm}      
\usepackage{algorithmic}    

\usepackage{graphicx}       
\usepackage{subcaption}     
\usepackage{float}          
\usepackage{booktabs}       
\usepackage{multirow}       
\usepackage{multicol}       
\usepackage{array}          
\usepackage{tabularx}       
\usepackage{makecell}       

\usepackage{xcolor}         
\usepackage{tikz}           
\usetikzlibrary{shapes, arrows, positioning, calc, fit, backgrounds}

\usepackage{listings}       
\usepackage{microtype}      
\usepackage{enumitem}       
\usepackage{setspace}       
\usepackage{balance}        

\usepackage{siunitx}        
\usepackage{textcomp}       
\usepackage{gensymb}        

\theoremstyle{plain}

\theoremstyle{definition}

\theoremstyle{remark}

\AtBeginDocument{%
  }

\copyrightyear{2026}
\acmYear{2026}
\setcopyright{cc}
\setcctype{by}
\acmConference[MobiCom '26]{The 32nd Annual International Conference on Mobile Computing and Networking}{October 26--30, 2026}{Austin, TX, USA}
\acmBooktitle{The 32nd Annual International Conference on Mobile Computing and Networking (MobiCom '26), October 26--30, 2026, Austin, TX, USA}
\acmDOI{10.1145/3795866.3844472}
\acmISBN{979-8-4007-2505-0/2026/10}

\acmSubmissionID{1307}

\begin{document}

\title{CRISP: Co-designing Pruning and Scheduling for Efficient MoE Inference on Edge Servers}

\author{Congwei Zhang}
\authornote{Congwei Zhang and Youming Tao contributed equally to this work.}
\email{xczhw@mail.sdu.edu.cn}
\affiliation{%
  \department{School of Computer Science and Technology}
  \institution{Shandong University}
  \city{Qingdao}
  \country{China}
}

\author{Youming Tao}
\authornotemark[1]
\email{tao@ccs-labs.org}
\affiliation{%
  \department{School of Electrical Engineering and Computer Science}
  \institution{TU Berlin}
  \city{Berlin}
  \country{Germany}
}

\author{Yifei Zou}
\authornote{Yifei Zou is the corresponding author.}
\email{yfzou@sdu.edu.cn}
\affiliation{%
  \department{School of Computer Science and Technology}
  \institution{Shandong University}
  \city{Qingdao}
  \country{China}
}

\author{Yu Yang}
\email{yuyang@cityu.edu.hk}
\affiliation{%
  \department{Department of Data Science}
  \institution{City University of Hong Kong}
  \city{Hong Kong}
  \country{China}
}

\author{Ruirui Zhang}
\email{sherryz@mail.sdu.edu.cn}
\affiliation{%
  \department{School of Computer Science and Technology}
  \institution{Shandong University}
  \city{Qingdao}
  \country{China}
}

\author{Xiaoyu Zhang}
\email{philzxy@mail.sdu.edu.cn}
\affiliation{%
  \department{School of Computer Science and Technology}
  \institution{Shandong University}
  \city{Qingdao}
  \country{China}
}

\author{Falko Dressler}
\email{dressler@ccs-labs.org}
\affiliation{%
  \department{School of Electrical Engineering and Computer Science}
  \institution{TU Berlin}
  \city{Berlin}
  \country{Germany}
}

\author{Xiuzhen Cheng}
\email{xzcheng@sdu.edu.cn}
\affiliation{%
  \department{School of Computer Science and Technology}
  \institution{Shandong University}
  \city{Qingdao}
  \country{China}
}

\author{Dongxiao Yu}
\email{dxyu@sdu.edu.cn}
\affiliation{%
  \department{School of Computer Science and Technology}
  \institution{Shandong University}
  \city{Qingdao}
  \country{China}
}

\renewcommand{\shortauthors}{Zhang et al.}

\begin{abstract}
Mixture-of-Experts (MoE) inference on memory-constrained edge platforms is often dominated by expert-weight movement rather than computation.
Although each token activates only a small subset of experts, routing is input-dependent, so all experts must remain accessible and memory movement dominates latency.
Existing approaches such as uniform pruning and expert dropping overlook two properties of MoE execution: expert combinations are highly skewed in frequency, and experts contribute asymmetrically within each combination.
We present CRISP, a combination-aware framework that co-designs pruning and runtime scheduling for efficient MoE serving on memory-constrained edge platforms.
CRISP reorders neurons by importance so one stored expert can support multiple execution widths, assigns differentiated execution to hot, warm, and cold combinations based on frequency and expert role, and exploits heterogeneous widths to overlap expert loading with computation.
Across Mixtral-8x7B and DeepSeek-MoE-16B, CRISP improves throughput by up to $2.1\times$ at the same memory budget while limiting accuracy loss to 3.6\% relative to full-width inference.
Our results highlight that efficient edge MoE serving requires jointly optimizing which expert capacity is preserved and how expert execution is scheduled.
\end{abstract}



\maketitle

\vspace{-4pt}
\section{Introduction}
Edge intelligence must operate within limited memory, computation, and communication budgets~\cite{chen2021distributed,tao2024private,xin2026sereno, zhang2024distributed}. Adapting model capacity and execution to these constraints is a recurring design goal across resource-aware learning and on-device inference~\cite{chen2025adaptive,tao2026autosub,yin2025elastic, taodistributed}. The growing demand for low-latency responses, local data processing, and offline availability has further motivated LLM deployment on edge computing platforms~\cite{yi2025edgemoe,song2025smallthinker, wu2025adaptive, zou2024distributed, zhang2026lisfc}.
However, the substantial parameter counts of modern models present two fundamental challenges: computational cost and memory footprint. Mixture-of-Experts (MoE) architectures~\cite{shazeer2017outrageously,jiang2024mixtral,fedus2022switch, zhang2026unified, zhang2025network} elegantly address the computational challenge through conditional computation, activating only a sparse subset of experts for each token. For instance, Mixtral-8×7B comprises 47B total parameters yet activates merely 13B per token, substantially reducing computational requirements while preserving model capacity.

The memory challenge, however, remains acute. Although only a few experts execute per token, all expert weights must remain accessible since routing decisions are input-dependent and cannot be determined a priori. When the complete model exceeds GPU memory capacity, a common scenario on edge platforms, inference systems typically employ hierarchical memory management, storing expert weights in host memory and transferring them to the GPU on demand. This offloading paradigm fundamentally shifts the performance bottleneck from computation to data movement. Empirical studies~\cite{moe-offload,yi2025edgemoe,wang2025d2moe, zhang2026balancing} demonstrate that I/O latency constitutes over 70\% of total inference time on bandwidth-constrained edge platforms. The resulting data-movement bottleneck is architecture-dependent: discrete-GPU systems incur explicit host-device transfers, whereas unified-memory platforms contend for shared memory bandwidth and cache capacity. In both cases, repeatedly accessing large expert weights leaves substantial execution time exposed to memory movement.

To mitigate the expert-weight I/O bottleneck, prior systems have primarily relied on runtime techniques such as offloading, caching, and prefetching~\cite{deepspeed_moe,tutel,powerinfer, wang2026distributed} to better manage or hide data movement. However, when expert-weight access substantially exceeds the available computation window, scheduling alone cannot eliminate the resulting stalls. Pruning provides a complementary opportunity by directly reducing the expert weights that must be accessed and computed for each routed token. More importantly, structured pruning exposes expert width as a controllable execution dimension, creating an opportunity to jointly optimize model capacity and runtime execution.

\begin{figure}[t]
    \centering
    \includegraphics[width=\linewidth]{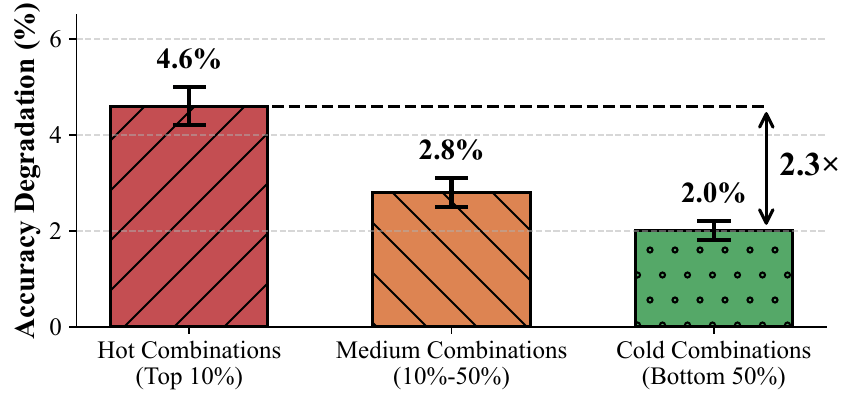}
    \caption{Uniform pruning disproportionately harms the most frequently used expert pairs.}
    \label{fig:uniform-pruning-limitation}
\end{figure}

\begin{figure}[t]
    \centering
    \includegraphics[width=\linewidth]{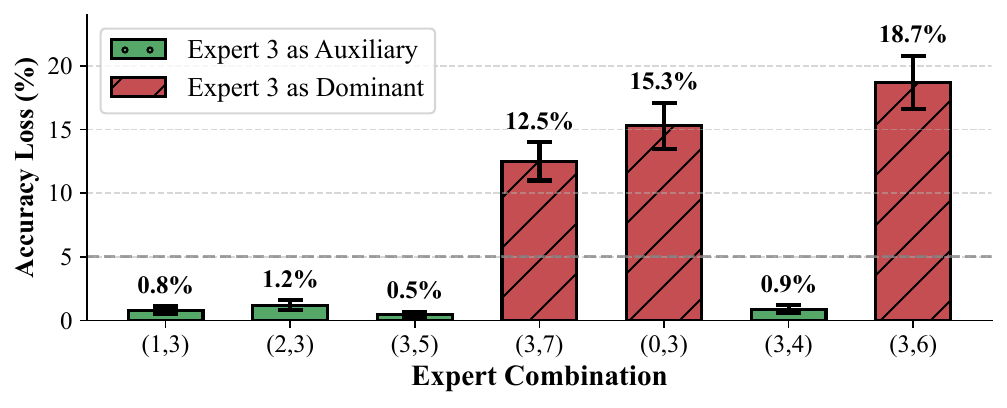}
    \caption{Expert dropping fails to capture the nuanced, combination-dependent roles of experts. This figure shows the accuracy loss when dropping Expert 3. Red bars indicate combinations where Expert 3 is a significant expert, while green bars indicate it is negligible.}
    \label{fig:expert-dropping-limitation}
\end{figure}

Existing MoE pruning methods, however, are poorly matched to actual expert usage. \textit{Uniform pruning}~\cite{pruning_magnitudes,wanda, chi2026efficient} applies the same pruning ratio across experts, ignoring the highly skewed distribution of expert combinations. As shown in Figure~\ref{fig:uniform-pruning-limitation}, under 50\% uniform pruning, frequently activated combinations suffer 2.3× greater accuracy degradation than rarely activated ones. \textit{Expert dropping}~\cite{expert_pruning_lu} instead removes selected experts entirely, but its binary granularity fails to capture combination-dependent expert roles: the same expert can be auxiliary in one combination yet dominant in another, as illustrated in Figure~\ref{fig:expert-dropping-limitation}. These observations suggest that pruning decisions should be made at the level of routed expert combinations, accounting for both combination frequency and expert role.

These limitations reveal a broader opportunity: existing pruning strategies are largely designed independently of runtime scheduling. Uniform pruning produces homogeneous execution widths, while expert dropping provides only binary keep-or-remove decisions. Neither exposes fine-grained, controllable width heterogeneity that can be systematically exploited at runtime. In contrast, assigning different execution widths according to combination frequency and expert role creates heterogeneous I/O and computation times, which can provide additional opportunities to overlap expert loading with computation. This suggests that pruning and runtime scheduling should be considered jointly rather than as independent optimization stages.


Realizing this combination-aware pruning-and-scheduling design requires jointly determining how much expert capacity to preserve, how to represent multiple execution widths efficiently, and how to schedule the resulting heterogeneous executions at runtime. This introduces three challenges.
\begin{itemize}[leftmargin=10pt,label=-]
    \item \textbf{Challenge~1 Storage efficiency.} Combination-aware pruning implies that the same expert may need to run at different widths in different combinations. A naive design would store a separate copy of the expert for each width, quickly eliminating the storage savings from pruning. The challenge is therefore to support flexible width selection at runtime without replicating expert weights.
    \item \textbf{Challenge~2 Combination strategy.} The number of possible expert combinations grows combinatorially with the number of experts. For a model with $n$ experts and top-$k$ routing, there are $\binom{n}{k}$ possible combinations. Differentiating among these combinations requires balancing accuracy, storage cost, and scheduling complexity. A fine-grained policy becomes difficult to manage at runtime, while a coarse one fails to capture meaningful differences in combination frequency and expert role.
    \item \textbf{Challenge~3 Scheduling coordination.} Differentiated pruning creates experts with different widths, which in turn leads to heterogeneous I/O and computation times. Wider experts take longer to load, but also longer to execute. This heterogeneity creates an opportunity to overlap expert loading with computation, but only if pruning and scheduling are designed together. Without such coordination, the latency benefits of differentiated pruning cannot be fully realized.
\end{itemize}

To address these challenges, we present \textit{CRISP} (\textit{Combination aware Reordering and I/O-Scheduled Pruning}), an edge MoE inference framework that co-designs expert representation, combination-aware pruning, and runtime scheduling to reduce expert-weight access.

\begin{itemize}[leftmargin=10pt,label=-]
    \item To address \textbf{Challenge~1 (storage efficiency)}, CRISP introduces importance-aware neuron reordering to support multiple execution widths from a single stored expert. CRISP ranks neurons using a calibration set and physically reorders each expert by importance, so that a width-$k$ execution simply accesses the first $k$ neurons. This avoids the storage redundancy of materializing separate weight copies for different widths while retaining flexible runtime pruning.
    \item To address \textbf{Challenge~2 (combination strategy)}, CRISP adopts a tiered execution policy based on combination frequency and expert role. Offline profiling partitions routed combinations into Hot, Warm, and Cold tiers and identifies the dominant and auxiliary experts within each combination. Hot combinations retain both experts at full width, Warm combinations reduce the auxiliary width, and Cold combinations skip the auxiliary expert. This policy preserves more capacity for frequently activated and dominant experts while pruning more aggressively where the impact is smaller.
    \item To address \textbf{Challenge~3 (scheduling coordination)}, CRISP introduces a pruning-aware I/O scheduler that exploits the execution heterogeneity created by differentiated pruning. It prioritizes wider experts, whose longer computation windows can overlap the loading of narrower experts. In this way, CRISP translates pruning-induced width heterogeneity into reduced exposed I/O latency.
\end{itemize}
The three mechanisms form a single co-design path rather than independent optimizations. Importance-aware reordering exposes multiple execution widths from one stored expert; tiered execution selects among these widths according to combination frequency and expert role; and the pruning-aware scheduler exploits the resulting width heterogeneity to overlap I/O with computation. Thus, the output of each component creates the capability required by the next: reordering makes differentiated pruning storage-efficient, differentiated pruning shapes the runtime I/O and computation profile, and scheduling translates this heterogeneity into latency reduction. We summarize our contributions as follows:

\begin{enumerate}[leftmargin=15pt]
    \item We identify two combination-level properties of MoE execution via empirical studies: expert combinations exhibit highly skewed activation frequencies, and co-activated experts contribute asymmetrically. These observations motivate pruning decisions based on both combination frequency and expert role.

    \item We develop importance-aware neuron reordering that enables multiple execution widths from a single stored expert through prefix truncation, avoiding the weight duplication that would otherwise arise from materializing multiple pruned variants.

    \item We co-design combination-aware differentiated pruning with runtime scheduling. A tiered policy assigns execution widths according to combination frequency and expert role, deliberately creating heterogeneous expert executions; a pruning-aware scheduler then exploits this heterogeneity to overlap expert loading with computation.

    \item We implement CRISP and evaluate it on Mixtral-8x7B and DeepSeek-MoE-16B. Under the same memory budget, CRISP improves throughput by up to 2.1$\times$ while limiting accuracy degradation to 3.6\% relative to full-width execution.
\end{enumerate}
\section{Empirical Observations}

Mixture-of-Experts (MoE) models route each token to a subset of specialized expert networks. A typical MoE layer consists of a gating network (router) and $n$ expert networks, where the router selects the top-$k$ experts based on learned routing scores. For instance, Mixtral-8x7B selects 2 out of 8 experts per token, activating 13B of its 47B total parameters.

On edge servers with limited GPU memory, expert weights must be offloaded to host memory and loaded on demand. Table~\ref{tab:io-breakdown} shows the I/O and computation breakdown for a single MoE layer on an NVIDIA RTX 4090 with PCIe 4.0. Loading two experts dominates the latency, motivating our focus on reducing expert size through pruning.

\begin{table}[t]
\centering
\caption{I/O and computation breakdown for a single MoE layer inference on NVIDIA RTX 4090 with PCIe 4.0 (Mixtral-8x7B, 4-bit quantization).}
\label{tab:io-breakdown}
\begin{tabular}{lcc}
\toprule
\textbf{Operation} & \textbf{Time (ms)} & \textbf{Percentage} \\
\midrule
Expert Loading (2 experts) & 11.2 & 78.0\% \\
Router Computation & 0.4 & 2.8\% \\
Expert Computation & 2.6 & 18.1\% \\
Other (memory copy, etc.) & 0.2 & 1.1\% \\
\midrule
\textbf{Total} & \textbf{14.4} & \textbf{100\%} \\
\bottomrule
\end{tabular}
\end{table}

The background analysis reveals that I/O bandwidth is the primary bottleneck for edge MoE inference, motivating model pruning as a direct solution. However, existing pruning methods apply uniform compression across all experts, overlooking the heterogeneous nature of expert usage patterns. In this section, we conduct empirical studies on MoE inference behavior to uncover optimization opportunities that inform the design of CRISP. All empirical studies in this section are conducted on Mixtral-8x7B. To ensure the representativeness of our observations, we use a calibration dataset consisting of 2,048 sequences randomly sampled from the C4 validation set~\cite{c4}. C4 is a massive, diverse corpus of web text widely used for LLM pre-training and evaluation, ensuring that the profiled activation patterns reflect general-purpose model usage rather than domain-specific idiosyncrasies.

\subsection{Power-law Distribution of Expert Combinations}
\begin{figure}[t]
    \centering
    \includegraphics[width=\columnwidth]{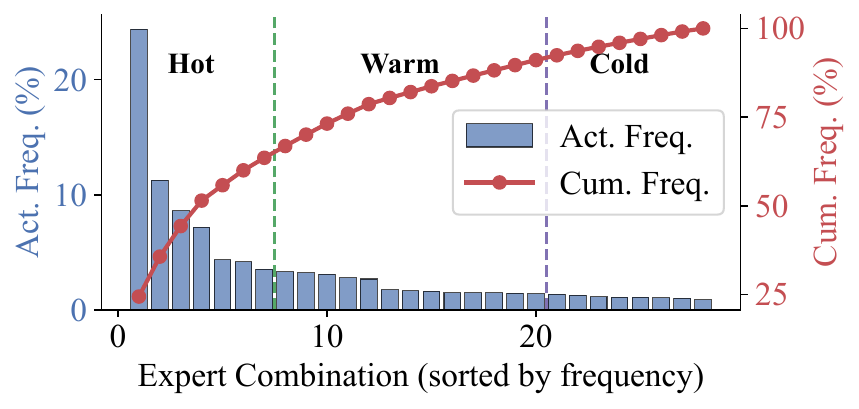}
    \caption{Distribution of expert combination activation frequency (Act. Freq.) and cumulative frequency (Cum. Freq.) in Mixtral-8x7B with top-2 routing, profiled on 2,048 sequences from the C4 validation set. A small fraction of combinations accounts for the majority of token activations, while most combinations are rarely used.}
    \label{fig:combination_distribution}
\end{figure}

In MoE models with top-2 routing, each token activates a pair of experts, resulting in $\binom{n}{2}$ possible combinations for a model with $n$ experts. We profile the activation frequency of all expert combinations across the C4 calibration dataset and present the results in Figure~\ref{fig:combination_distribution}. The distribution exhibits a clear power-law pattern: a small fraction of expert combinations handle the vast majority of tokens, while most combinations are rarely activated. Specifically, the top 25\% of combinations account for over 60\% of all token activations, while the bottom 29\% of combinations collectively handle less than 10\% of tokens. This skewed distribution is consistent across different layers of the model, though the specific hot combinations vary by layer.

\textit{Summary:} The highly imbalanced activation pattern suggests that uniform pruning is suboptimal---frequently activated combinations deserve more capacity to preserve accuracy, while rarely activated combinations can tolerate aggressive compression with minimal impact on overall model quality.

\subsection{Asymmetric Expert Roles in Combinations}
\begin{figure}[t]
    \centering
    \includegraphics[width=\columnwidth]{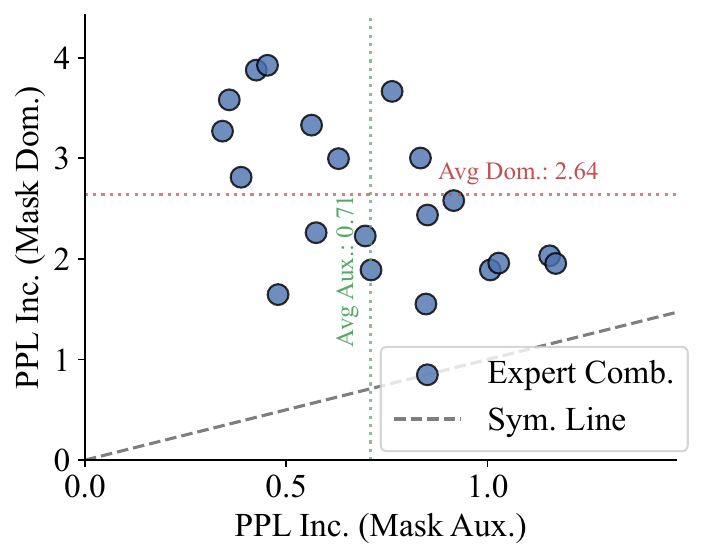}
    \vspace{-25pt}
    \caption{Asymmetric roles of experts within combinations, visualized by perplexity increase (PPL Inc.) when masking each expert in the top 20 most frequent combinations. Most points deviate from the diagonal (Sym. Line), indicating that one dominant expert contributes significantly more than its auxiliary counterpart.}
    \vspace{-10pt}
    \label{fig:expert_asymmetry}
\end{figure}

Even within the same expert combination, the two activated experts do not contribute equally to the output. To quantify this asymmetry, we conduct an ablation study where we selectively mask one expert in each combination (replacing its output with zeros) and measure the resulting increase in perplexity. Figure~\ref{fig:expert_asymmetry} shows the results for the top 20 most frequent combinations. For each combination, we plot the perplexity increase caused by masking each of the two experts. If both experts were equally important, the points would cluster along the diagonal; instead, we observe significant deviation, indicating that one expert typically has a much larger impact than the other.

We refer to the expert whose removal causes greater accuracy degradation as the \textit{dominant expert}, and the other as the \textit{auxiliary expert}. Across all combinations, masking the dominant expert increases perplexity by an average of 2.64, while masking the auxiliary expert increases perplexity by only 0.71---a 3.7$\times$ difference. This asymmetry suggests functional specialization: the dominant expert captures the primary functionality required for the input, while the auxiliary expert provides supplementary refinement.

\textit{Summary:} The asymmetric roles of experts within combinations open an opportunity for fine-grained differentiated pruning. Rather than treating both experts identically, we can preserve full capacity for dominant experts while aggressively pruning auxiliary experts, achieving memory savings with minimal accuracy loss.

\subsection{Opportunity for I/O-Computation Overlap}
\begin{figure}[t]
    \centering
    \includegraphics[width=\columnwidth]{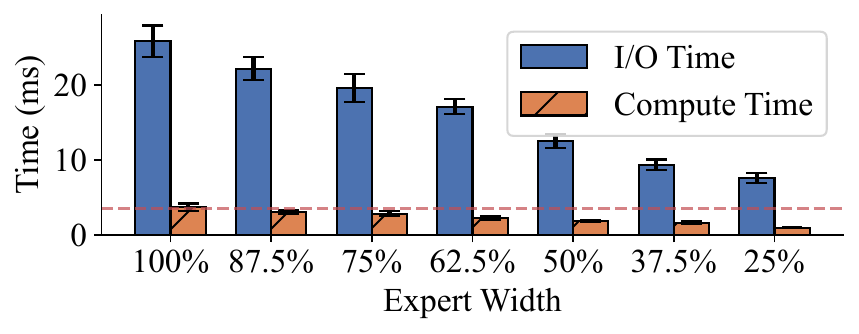}
    \caption{I/O and computation time of a single expert under different pruning ratios. As expert width decreases, both I/O and computation time are reduced, but at different rates, creating opportunities to overlap the I/O of narrow experts with the computation of wider experts.}
    \label{fig:io_compute_ratio}
\end{figure}

Differentiated pruning produces experts of varying widths, which naturally leads to heterogeneous I/O and computation times. We measure the I/O time (loading from host memory via PCIe 4.0) and computation time (GPU execution) for a single expert at different pruning ratios. As shown in Figure~\ref{fig:io_compute_ratio}, both I/O and computation time decrease as the expert width is reduced, but at different rates. A full-width expert requires 7.5 ms for I/O and 0.8 ms for computation, while a 50\%-width expert requires 3.8 ms for I/O and 0.4 ms for computation.

This heterogeneity creates an opportunity for strategic scheduling. When executing a sequence of experts with different widths, the computation of a wider expert can potentially mask the I/O of subsequent narrower experts. In contrast, uniform pruning produces homogeneous experts that offer limited room for such overlap. To quantify this opportunity, we simulate a naive sequential execution (load-then-compute for each expert) versus an ideal overlapped execution for a typical MoE layer with our tiered pruning policy. The naive approach results in 90\% I/O-computation bubble, while perfect overlap could reduce this to 38\%.

\textit{Summary:} The heterogeneous expert sizes resulting from differentiated pruning are not merely a byproduct but an opportunity. A pruning-aware scheduler that exploits this heterogeneity can significantly reduce I/O stalls, translating the memory savings from pruning into actual latency improvements.
\section{System Design}

\subsection{System Overview}

\begin{figure}[h]
    \centering
    \includegraphics[width=.95\columnwidth]{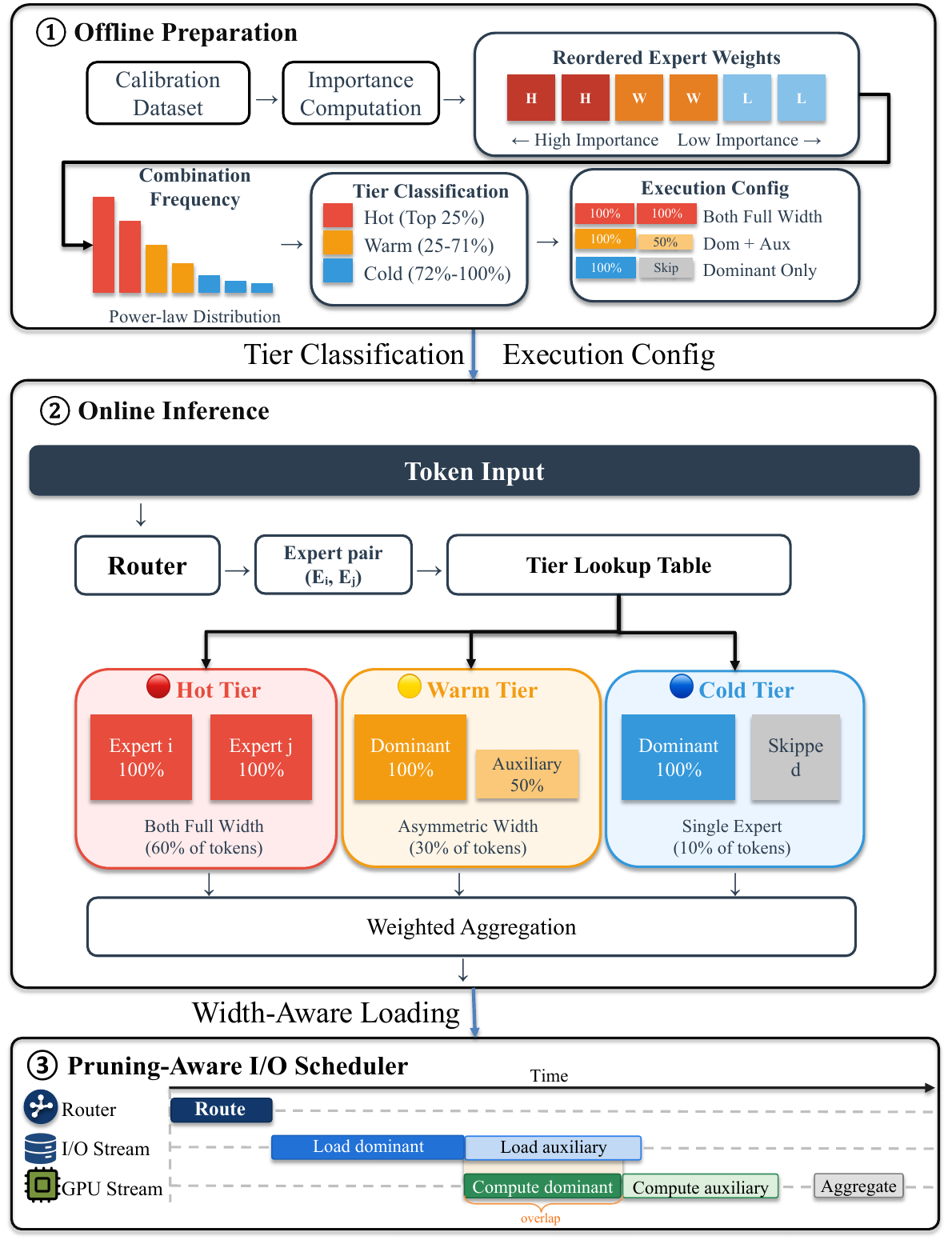}
    \caption{Overall architecture of CRISP. The offline phase profiles expert combinations and reorders expert weights to produce tier-aware execution metadata and width-flexible experts. During inference, the router activates an expert combination whose tier determines the execution width of each expert. A pruning-aware scheduler prioritizes wider experts so that their longer computation overlaps with loading of narrower experts, hiding I/O latency.}
    \label{fig:system_overview}
    \vspace{-10pt}
\end{figure}

Figure~\ref{fig:system_overview} illustrates the overall architecture of CRISP, which consists of an offline preparation phase and an online inference phase.

In the offline phase, CRISP performs three key preparation steps. First, it profiles expert combination activation frequencies on a calibration dataset (empirically verified to be representative of the target workload) to classify combinations into Hot, Warm, and Cold tiers. Second, it computes neuron importance scores for each expert based on activation magnitudes and identifies the dominant expert within each combination. Third, it reorders the weight matrices of each expert according to neuron importance, producing a single stored copy that supports arbitrary-width truncation.

CRISP's calibration-derived neuron rankings, combination frequencies, and expert roles depend on both the model parameters and the deployment workload. Changes may arise from shifts in the input distribution or from model updates through fine-tuning~\cite{chen2026fedsubmuon,zhang2025train}, unlearning~\cite{tao2024communication,li2026not}, and model editing~\cite{tao2026one-shot}. The corresponding profiling artifacts should therefore be reassessed when the model or workload changes substantially. For a workload change, CRISP collects a new calibration sample and refreshes the combination statistics, tier assignments, and expert roles. If the expert weights are updated, neuron importance scores and the associated weight ordering may also need to be recomputed. These changes affect offline preparation while retaining the same online execution mechanism.

In the online phase, CRISP executes inference through coordinated execution. When a token arrives at an MoE layer, the router first determines which expert combination to activate. CRISP then consults the precomputed tier assignment to determine the execution configuration: full-width for both experts (Hot), full-width for dominant and reduced-width for auxiliary (Warm), or dominant expert only (Cold). Meanwhile, the pruning-aware scheduler coordinates the already-routed expert tasks, overlapping the computation of one selected expert with the I/O loading of another, prioritizing wider experts to maximize overlap efficiency.

\noindent\textbf{Design rationale.} The three components of CRISP address interdependent decisions and are insufficient when applied independently. Importance-aware reordering makes multiple execution widths available from one stored expert, but it does not determine which width should be selected for a given combination. Conversely, tiered pruning determines combination-specific widths, but without reordering it would require separate materialized copies for different widths, offsetting its storage benefit. Scheduling alone can only reorder the available I/O and computation tasks; it cannot create the heterogeneous execution times that provide CRISP with additional overlap opportunities. CRISP therefore couples the three components so that reordering enables flexible widths, tiering assigns those widths, and scheduling exploits the resulting heterogeneity.

\subsection{Importance-Aware Neuron Reordering}

A key enabler of differentiated pruning is the ability to execute experts at varying widths without storing multiple copies. This subsection presents our importance-aware neuron reordering mechanism that achieves this goal through a single reorganized weight representation.

\subsubsection{Problem Formulation.}
Consider a standard MoE expert implemented as a feed-forward network (FFN) with a gated architecture. Each expert consists of three weight matrices: the gate projection $W_{\text{gate}} \in \mathbb{R}^{d \times d_{\text{ff}}}$, the up projection $W_{\text{up}} \in \mathbb{R}^{d \times d_{\text{ff}}}$, and the down projection $W_{\text{down}} \in \mathbb{R}^{d_{\text{ff}} \times d}$, where $d$ is the hidden dimension and $d_{\text{ff}}$ is the intermediate dimension. The expert computation follows:
\begin{equation}
\text{FFN}(x) = \left( \sigma(xW_{\text{gate}}) \odot xW_{\text{up}} \right) W_{\text{down}}
\end{equation}
where $\sigma$ denotes the activation function (e.g., SiLU) and $\odot$ denotes element-wise multiplication. To prune an expert to width $k < d_{\text{ff}}$, we need to select $k$ neurons (columns of $W_{\text{gate}}$ and $W_{\text{up}}$, rows of $W_{\text{down}}$) to retain. The naive approach would require storing separate weight matrices for each desired width, leading to storage overhead proportional to the number of supported widths.

\subsubsection{Neuron Importance Metric.}
We adopt an activation-magnitude-based importance metric that measures each neuron's contribution to the expert output. Given a calibration dataset $\mathcal{D}$, we compute the importance score of the $i$-th neuron as the average magnitude of its activation across all samples:
\begin{equation}
s_i = \frac{1}{|\mathcal{D}|} \sum_{x \in \mathcal{D}} \left| \sigma(xW_{\text{gate}})_i \cdot (xW_{\text{up}})_i \right|
\end{equation}
where the subscript $i$ denotes the $i$-th element of the vector. This metric captures the effective contribution of each neuron after the gating mechanism, reflecting both the gate activation and the up-projection magnitude.

\subsubsection{Weight Reordering and Runtime Truncation.}
Once importance scores are computed, we sort neurons in descending order of importance and physically reorder the weight matrices accordingly. Let $\pi$ denote the permutation that sorts neurons by importance, i.e., $s_{\pi(1)} \geq s_{\pi(2)} \geq \cdots \geq s_{\pi(d_{\text{ff}})}$. We construct reordered weights as:
\begin{equation}
\tilde{W}_{\text{gate}} = W_{\text{gate}}[:, \pi], \quad \tilde{W}_{\text{up}} = W_{\text{up}}[:, \pi], \quad \tilde{W}_{\text{down}} = W_{\text{down}}[\pi, :]
\end{equation}
where $[:, \pi]$ denotes column permutation and $[\pi, :]$ denotes row permutation. This reordering preserves mathematical equivalence: $\text{FFN}(x)$ computed with reordered weights produces identical outputs to the original. The key benefit is that truncating to the first $k$ neurons now yields the optimal width-$k$ sub-expert, as these neurons have the highest importance scores. At inference time, executing an expert at width $k$ requires only loading the first $k$ columns of $\tilde{W}_{\text{gate}}$ and $\tilde{W}_{\text{up}}$, and the first $k$ rows of $\tilde{W}_{\text{down}}$. This is achieved through simple offset-based memory access without any runtime computation for neuron selection, and the I/O volume scales linearly with the target width.

\subsection{Hotness-Based Tiered Execution Policy}

With the reordering mechanism enabling flexible width selection, we now address how to assign appropriate widths to different expert combinations. This subsection presents our hotness-based tiered execution policy that exploits the power-law distribution of combination frequencies and the asymmetric roles of experts within combinations.

\subsubsection{Combination Tier Classification.}
Expert combinations exhibit highly skewed activation frequencies, as established in our motivation analysis. We leverage this property by classifying combinations into three tiers with differentiated execution strategies. Given the activation frequency $f_c$ for each combination $c$, we sort all combinations by frequency and partition them according to cumulative frequency thresholds. \emph{Hot tier} contains combinations whose cumulative frequency accounts for the top $\alpha$ of total activations; these are accessed frequently and have the highest impact on overall model accuracy. \emph{Warm tier} contains combinations whose cumulative frequency falls between $\alpha$ and $\beta$ of total activations, representing moderately frequent access patterns. \emph{Cold tier} contains combinations in the bottom $1 - \beta$ of total activations, which are rarely activated. In our implementation, we set $\alpha = 0.6$ and $\beta = 0.9$, resulting in approximately 25\% of combinations classified as Hot, 46\% as Warm, and 29\% as Cold. We choose $\alpha$ and $\beta$ based on the measured activation-frequency distribution on the calibration set to balance accuracy preservation (by capturing the majority of activations in the Hot/Warm tiers) and I/O reduction (by allocating a sufficiently large tail to the Cold tier), and keep them fixed across all experiments.

\subsubsection{Dominant Expert Identification.}
Experts within a combination play asymmetric roles, as shown in our motivation analysis. For each combination, we identify the dominant expert through an offline ablation process. Specifically, for combination $(e_i, e_j)$, we measure the perplexity increase when masking each expert individually: $\Delta_i = \text{PPL}(\text{mask } e_i) - \text{PPL}(\text{original})$. The expert with the larger $\Delta$ value is designated as the dominant expert, as its removal causes greater accuracy degradation. This classification is performed once during offline preparation and stored in a lookup table indexed by combination identity.

\subsubsection{Tiered Execution Strategies.}
Each tier employs a distinct execution strategy that balances accuracy preservation with I/O efficiency. For \emph{Hot tier} combinations, both experts execute at full width $d_{\text{ff}}$, preserving maximum model capacity for the combinations that process the majority of tokens and ensuring minimal accuracy degradation on common inputs. For \emph{Warm tier} combinations, the dominant expert executes at full width while the auxiliary expert executes at reduced width $k_{\text{warm}} = 0.5 \cdot d_{\text{ff}}$. This strategy exploits role asymmetry: the dominant expert maintains full capacity to capture primary functionality, while the auxiliary expert provides partial refinement at lower cost. For \emph{Cold tier} combinations, only the dominant expert executes at full width, and the auxiliary expert is skipped entirely. This aggressive strategy is justified by the low activation frequency limiting impact on overall accuracy, and the functional overlap between co-activated experts allowing the dominant expert to partially compensate for the missing auxiliary.

\subsubsection{Memory Footprint Analysis.}
The tiered policy significantly reduces the effective memory footprint compared to full-width execution. Let $p_H$, $p_W$, and $p_C$ denote the fraction of tokens processed by Hot, Warm, and Cold combinations respectively. The average memory access per token is:
\begin{equation}
M_{\text{avg}} = p_H \cdot 2M_{\text{full}} + p_W \cdot (M_{\text{full}} + M_{\text{narrow}}) + p_C \cdot M_{\text{full}}
\end{equation}
where $M_{\text{full}}$ and $M_{\text{narrow}}$ denote the memory footprint of a full-width and narrow-width expert respectively. With typical values of $p_H = 0.60$, $p_W = 0.30$, $p_C = 0.10$, and $k_{\text{warm}} = 0.5 \cdot d_{\text{ff}}$, this yields an average reduction of 13\% in per-token memory access compared to the baseline.

\subsection{Pruning-Aware I/O Scheduler}

The tiered execution policy produces experts of heterogeneous widths, creating opportunities for I/O-computation overlap. This subsection presents our pruning-aware scheduler that exploits this heterogeneity through dependency-aware scheduling after routing.

\subsubsection{Dependency-Preserving Scheduling Architecture.}
In MoE inference with expert offloading, each layer involves two phases: I/O (loading activated experts from host memory to GPU) and computation (executing the experts on GPU). A naive sequential execution loads and computes each selected expert in turn, leaving significant idle time on both the I/O channel and the GPU. Once the current layer's router returns the selected experts and their execution widths, CRISP places these known tasks into the pending and ready queues. While the GPU executes a ready expert, the separate I/O stream loads another selected expert in parallel. After all selected experts finish and their outputs are aggregated, model execution proceeds to the next layer in dependency order.

\subsubsection{Width-Aware Scheduling Strategy.}
Within this scheduler, the order in which we load and execute experts significantly impacts overlap efficiency. Our key insight is that wider experts have longer computation times, which can mask more I/O latency for subsequent experts. Based on this observation, we adopt the following scheduling strategy: when multiple experts are pending for execution, we prioritize the expert with larger width. This ensures that wider experts begin computation earlier, providing longer windows to overlap with subsequent I/O operations.

Consider a Warm-tier combination with dominant expert $e_d$ (full-width) and auxiliary expert $e_a$ (narrow-width). Our scheduler loads and executes $e_d$ first. While $e_d$ computes, we load $e_a$ on the separate I/O stream. When $e_d$ completes, $e_a$ is already resident in GPU memory and can execute immediately. Algorithm~\ref{alg:scheduler} presents the detailed scheduling procedure. The scheduler maintains two queues: a pending queue of experts awaiting I/O and a ready queue of experts loaded and awaiting computation. At each scheduling decision point, it selects the widest expert from the ready queue for computation while simultaneously initiating I/O for the highest-priority pending expert.

\begin{algorithm}[t]
\caption{Pruning-Aware I/O Scheduling}
\label{alg:scheduler}
\begin{algorithmic}[1]
\REQUIRE Token input $x$, MoE layers $\{L_1, \ldots, L_N\}$
\STATE Initialize $\text{pending} \leftarrow \emptyset$, $\text{ready} \leftarrow \emptyset$
\STATE Execute router of $L_1$, get experts $E_1$ with widths $W_1$
\STATE Add $(E_1, W_1)$ to pending sorted by width descending
\FOR{$l = 1$ to $N$}
    \WHILE{pending for layer $l$ is not empty \OR ready is not empty}
        \IF{ready is not empty}
            \STATE $e \leftarrow$ pop widest expert from ready
            \STATE \textbf{async} start computation of $e$
        \ENDIF
        \IF{pending is not empty \AND I/O channel is idle}
            \STATE $e' \leftarrow$ pop highest-priority expert from pending
            \STATE \textbf{async} start loading $e'$, add to ready when done
        \ENDIF
    \ENDWHILE
    \STATE Wait for expert computations of layer $l$ and aggregate their outputs
    \IF{$l < N$}
        \STATE Continue dependency-ordered execution to the router of $L_{l+1}$
        \STATE Get experts $E_{l+1}$ with widths $W_{l+1}$ and add them to pending
    \ENDIF
\ENDFOR
\end{algorithmic}
\end{algorithm}

\subsubsection{Overlap Efficiency Analysis.}
The effectiveness of our scheduling strategy depends on the ratio between computation time and I/O time for experts of different widths. Let $T_{\text{io}}(w)$ and $T_{\text{comp}}(w)$ denote the I/O and computation time for an expert of width $w$. For our scheduling to fully hide the I/O of a narrow expert $e_a$ behind the computation of a wide expert $e_d$, we require $T_{\text{comp}}(w_d) \geq T_{\text{io}}(w_a)$. Based on our profiling results, a full-width expert has $T_{\text{comp}} = 0.8$ ms, while a 50\%-width expert has $T_{\text{io}} = 3.8$ ms. Although a single expert's computation cannot fully mask the I/O of another expert, the scheduler uses the available computation window of the full-width expert to overlap part of the narrow expert's transfer. Under batched execution, expert tasks from other requests that have completed routing can provide additional ready and pending work. This confirms that our scheduling strategy can effectively hide a significant portion of the I/O latency of narrow auxiliary experts behind the computation of full-width dominant experts in Warm-tier combinations.
\section{Implementation}
We have implemented a fully functional prototype of CRISP with approximately 5,000 lines of code in Python and CUDA, built on top of PyTorch. The source code will be made publicly available. This section describes the key implementation details across offline preparation and online inference.

\subsection{Offline Preparation Pipeline}

The offline phase preprocesses the model to enable efficient online inference. We implement this as a three-stage pipeline that can be executed once and reused across deployments.

\begin{itemize}[leftmargin=10pt,label=-]
    \item \textbf{Stage 1 (Combination Profiling).} We run the target MoE model on a calibration dataset (we use 2048 samples from C4 in our experiments) and record the expert combinations activated by each token. For each MoE layer, we maintain a frequency counter for all $\binom{n}{2}$ possible combinations, where $n$ is the number of experts. We then sort combinations by frequency and apply cumulative thresholds to partition them into Hot, Warm, and Cold tiers. The resulting tier assignments are stored as a lookup table indexed by the combination identifier $(e_i, e_j)$ where $e_i < e_j$.
    \item \textbf{Stage 2 (Importance Computation).} For each expert, we compute neuron importance scores by accumulating activation magnitudes across the calibration dataset. Specifically, we instrument the forward pass to capture the intermediate activations after the gating mechanism, compute element-wise products between gate and up projections, and maintain running averages for each neuron. To identify dominant experts within each combination, we perform pairwise ablation studies: for each combination in the Warm and Cold tiers, we measure perplexity degradation when masking each expert individually and designate the expert with larger degradation as dominant.
    \item \textbf{Stage 3 (Weight Reordering).} Based on the computed importance scores, we generate a permutation index for each expert that sorts neurons in descending order of importance. We then apply this permutation to physically reorder the weight matrices ($W_{\text{gate}}$, $W_{\text{up}}$, $W_{\text{down}}$) and save the reordered weights to disk. The reordered model maintains the same file format as the original, ensuring compatibility with standard model loading utilities.
\end{itemize}

The entire offline pipeline completes in approximately 1 hour for Mixtral-8x7B on a single NVIDIA RTX 4090 GPU, and the resulting artifacts (tier assignments, dominant expert mappings, reordered weights) add negligible storage overhead beyond the original model size.

\subsection{Online Inference Engine}

The online inference engine executes the prepared model with our tiered execution policy and pruning-aware scheduling. We implement this atop PyTorch with custom CUDA kernels for performance-critical operations.

\subsubsection{Expert Loading} We implement width-aware expert loading that transfers only the required portion of each expert's weights from host memory to GPU memory. For an expert at width $k$, we load the first $k$ columns of $W_{\text{gate}}$ and $W_{\text{up}}$, and the first $k$ rows of $W_{\text{down}}$. This is achieved through strided memory copies that skip over the pruned neurons, avoiding the need to load and then discard unnecessary data. We use CUDA streams to enable asynchronous transfers that can overlap with GPU computation.

\subsubsection{Truncated Expert Computation} We develop custom CUDA kernels for computing truncated expert outputs. Given an input tensor and the loaded partial weights, our kernels perform the gated FFN computation using only the first $k$ neurons. We optimize these kernels for the specific width configurations used in our tiered policy (full-width, 50\%-width) to maximize GPU utilization. For the Cold tier where only one expert executes, we skip the second expert entirely without allocating any GPU memory for it.

\subsubsection{Dependency-Preserving Scheduling.} We implement the pruning-aware scheduler using PyTorch's CUDA stream mechanism. We maintain separate streams for I/O transfers and GPU computation, with explicit synchronization points to ensure correctness. The scheduler maintains priority queues for pending and ready experts, making scheduling decisions based on expert widths as described in Algorithm~\ref{alg:scheduler}. Expert tasks are inserted into these queues after the corresponding current-layer router returns. The computation stream executes ready experts while the I/O stream loads other selected experts, and model execution advances only after the current expert outputs are aggregated.

\subsubsection{Tier Lookup.} At runtime, determining the execution configuration for an activated combination requires a single lookup into the precomputed tier assignment table. We implement this table as a GPU-resident tensor for fast access, with the combination identifier $(e_i, e_j)$ mapped to a linear index. The lookup returns the tier classification and, for Warm and Cold tiers, the identity of the dominant expert.

\subsection{Integration and Portability}

Our implementation is designed as a modular extension that can integrate with existing inference frameworks. The core components (weight reordering, tiered execution, pruning-aware scheduling) are implemented as drop-in replacements for standard MoE layer execution. We provide adapter interfaces for integration with popular frameworks including Hugging Face Transformers and vLLM. The width-aware loading and truncated computation kernels are implemented using NVIDIA's CUTLASS library, targeting Ampere and Ada Lovelace architectures, but the algorithmic approach is architecture-agnostic and can be ported to other platforms.

\section{Evaluation}

\subsection{Experimental Setup}
\begin{figure}[t]
\centering
\includegraphics[width=\columnwidth]{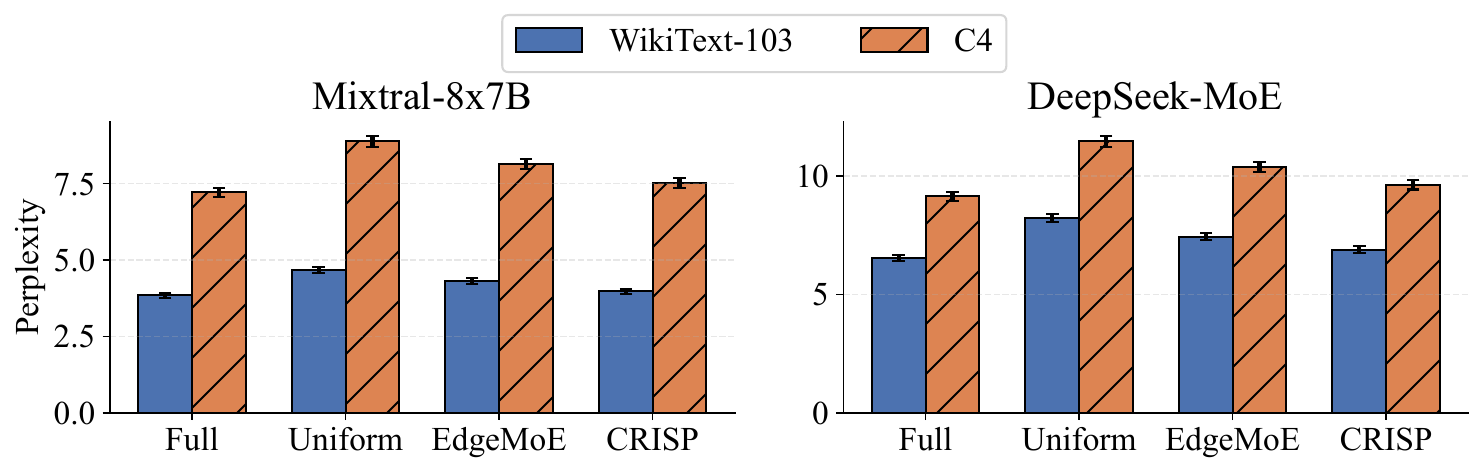}
\caption{Perplexity comparison on WikiText-103 and C4 datasets (lower is better).}
\label{fig:perplexity}
\end{figure}

\begin{figure}[t]
\centering
\includegraphics[width=\columnwidth]{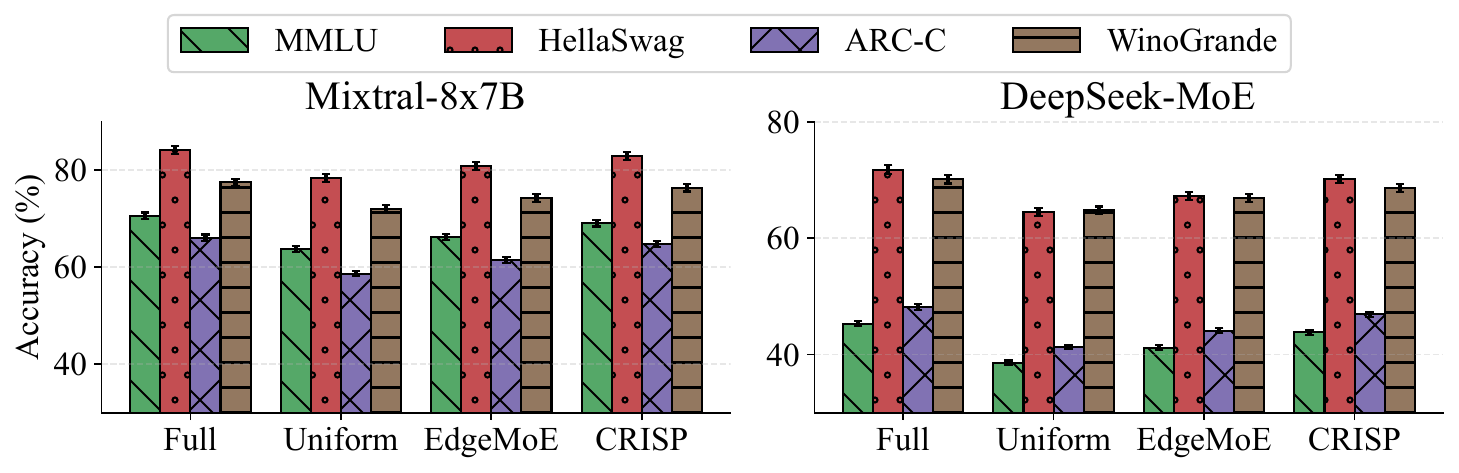}
\caption{Zero-shot accuracy on downstream tasks (higher is better).}
\label{fig:accuracy}
\end{figure}

\subsubsection{Models.} We evaluate CRISP on two representative MoE models: Mixtral-8x7B~\cite{mixtral} and DeepSeek-MoE-16B~\cite{deepseek-moe}. Mixtral-8x7B consists of 32 MoE layers with 8 experts each, using top-2 routing, resulting in a total of 46.7B parameters with 12.9B activated per token. DeepSeek-MoE-16B employs a finer-grained expert design with 64 experts per layer and top-6 routing, totaling 16.4B parameters with 2.8B activated per token.

\subsubsection{Hardware Platforms.} We conduct experiments on an embedded mobile edge platform and a controlled edge-gateway testbed. We use an NVIDIA Jetson AGX Orin with 32GB unified memory and 204 GB/s memory bandwidth to represent a typical mobile/embedded deployment. We additionally use a desktop-class GPU (NVIDIA RTX 4090, 24GB VRAM) paired with 64GB host memory over PCIe 4.0 x16 (32 GB/s bandwidth) as a controllable testbed to isolate the offloading bottlenecks under different memory constraints. To reflect smaller mobile GPUs, we cap the effective GPU memory available for expert caching to be insufficient to hold all experts (down to as low as 10\% of total expert size) on both platforms, thereby forcing expert offloading. The two systems expose different forms of data-movement cost: explicit host-to-device transfers on the RTX 4090 and contention for shared memory and caches on the Jetson Orin.

\subsubsection{Baselines.} We compare CRISP against the following methods: (1) \textit{Full Offloading} (FO), which loads complete experts on-demand without any pruning; (2) \textit{Uniform Pruning} (UP), which applies the same pruning ratio to all experts regardless of activation patterns; (3) \textit{MoE-Offload} (MO)~\cite{moe-offload}, a state-of-the-art offloading system that optimizes I/O scheduling but does not exploit differentiated pruning; and (4) \textit{EdgeMoE} (EM)~\cite{edgemoe}, which employs expert-wise adaptive precision but treats all activations uniformly.

\subsubsection{Benchmarks.} For accuracy evaluation, we measure perplexity on WikiText-103~\cite{wikitext} and C4~\cite{c4} datasets, and report accuracy on downstream tasks including MMLU~\cite{mmlu}, HellaSwag~\cite{hellaswag}, ARC-Challenge~\cite{arc}, and WinoGrande~\cite{winogrande}. For throughput evaluation, we measure token generation throughput (tokens/second) using input sequences of varying lengths sampled from the ShareGPT dataset~\cite{sharegpt}. We also integrate platform power telemetry over each generation run and divide by the number of generated tokens to report energy per token.

\subsubsection{Tier Thresholds.} We set the cumulative-frequency thresholds for combination tiering to $\alpha = 0.6$ and $\beta = 0.9$ based on a pilot profiling run on the calibration set, and keep them fixed across all experiments. Here, $\alpha$ and $\beta$ are cutoffs on the cumulative activation mass after combinations are sorted by frequency; they do not denote fixed expert identities.

\subsection{Accuracy Evaluation}
\label{subsec:accuracy}

\subsubsection{Model Accuracy Preservation.} We first evaluate whether CRISP preserves model accuracy under its differentiated pruning strategy. Figure~\ref{fig:perplexity} reports the perplexity results, and Figure~\ref{fig:accuracy} presents the downstream task accuracy for both models.
On Mixtral-8x7B, CRISP achieves a perplexity of \textbf{3.98} on WikiText-103, representing only a \textbf{3.6}\% increase compared to full-precision execution (3.84). In contrast, UP at a comparable memory reduction level degrades perplexity to \textbf{4.67}, a \textbf{21.6}\% increase. This demonstrates the effectiveness of our hotness-based tiered policy: by preserving full precision for frequently activated combinations while aggressively pruning cold combinations, CRISP maintains accuracy on common inputs that dominate overall perplexity.
For downstream tasks, CRISP preserves \textbf{97.4}\% of the original accuracy on average across all benchmarks, while UP retains only \textbf{90.6}\%. The accuracy gap is most pronounced on knowledge-intensive tasks like MMLU, where CRISP achieves \textbf{69.1}\% compared to \textbf{63.8}\% for UP. This suggests that our dominant expert identification successfully preserves the critical expert for each combination.
DeepSeek-MoE exhibits similar trends, with CRISP achieving \textbf{6.89} perplexity versus \textbf{8.21} for UP. The finer-grained expert structure of DeepSeek-MoE (64 experts with top-6 routing) creates more diverse combination patterns, which our tiered policy effectively exploits.

\subsubsection{Iso-Memory Comparison.}
\begin{figure}[t]
    \centering
    \includegraphics[width=\linewidth]{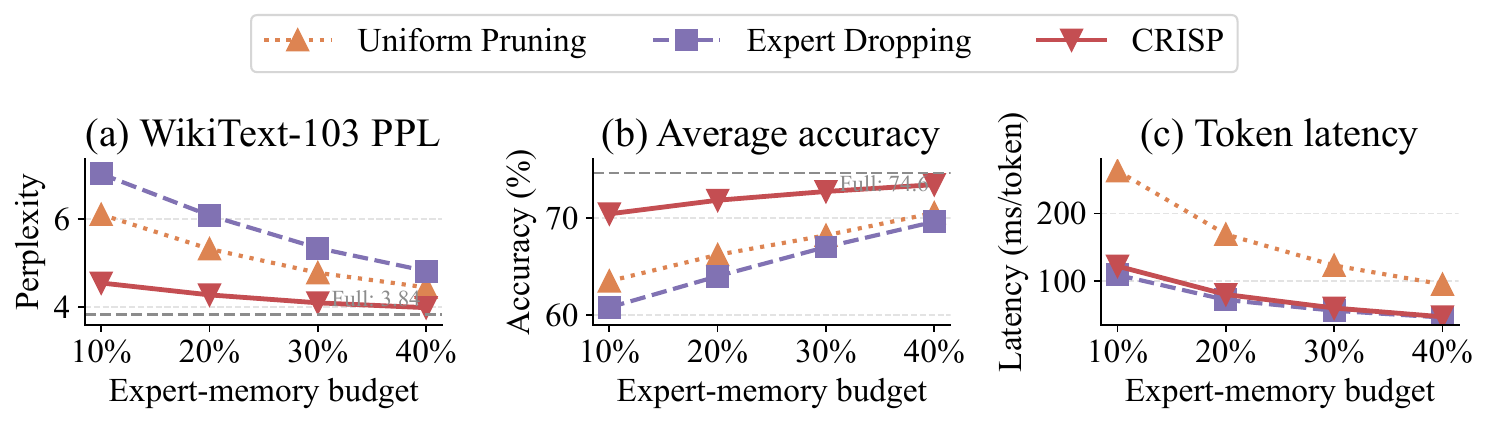}
    \caption{Iso-memory comparison of CRISP with Uniform Pruning and Expert Dropping under matched expert-memory budgets. Lower perplexity and latency are better, while higher average accuracy is better. Dashed horizontal lines show the quality of full-width execution.}
    \label{fig:iso_memory_accuracy}
\end{figure}

To compare quality under the same resource constraint, we evaluate CRISP against Uniform Pruning and Expert Dropping at matched expert-memory budgets. Figure~\ref{fig:iso_memory_accuracy} reports WikiText-103 perplexity, average downstream accuracy, and token latency as the available expert-memory budget varies from 10\% to 40\%. Across all evaluated budgets, CRISP preserves lower perplexity and higher average accuracy than both pruning baselines. At a 40\% expert-memory budget, for example, CRISP achieves 73.4\% average accuracy, compared with 70.5\% for Uniform Pruning and 69.6\% for Expert Dropping. Its 47 ms/token latency is close to Expert Dropping (46 ms/token) and substantially lower than Uniform Pruning (95 ms/token), showing that CRISP provides a more favorable quality-latency trade-off under the same memory constraint.

\subsubsection{Tier-conditioned accuracy.} Hot, Warm, and Cold routing events retain 99.9\%, 99.1\%, and 94.7\% of full-width accuracy, respectively. Cold combinations account for only 10\% of routing events; within this tier, skipping the auxiliary expert reduces I/O and latency at the cost of a larger conditional accuracy drop. Detailed Cold-tier trade-offs are reported in Appendix~\ref{app:tier_conditioned_accuracy}.

\subsubsection{Robustness to workload shift.} We further evaluate cross-domain transfer of profiling metadata among MMLU, HellaSwag, and WinoGrande. Directly transferring metadata decreases target-workload accuracy by at most 2.7 percentage points, while refreshing the metadata on the target workload recovers accuracy to within 0.5 points of the target-matched setting with similar throughput. Detailed results are provided in Appendix~\ref{app:workload_shift}.


\subsection{Throughput Evaluation}
\label{subsec:throughput}

\begin{figure}[t]
    \centering
    \includegraphics[width=.9\columnwidth]{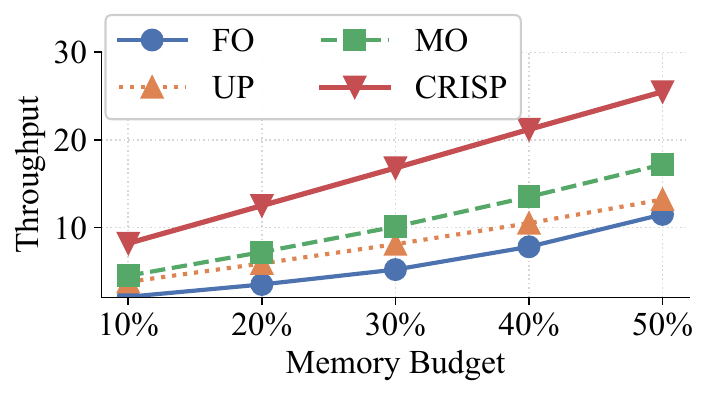}
    \caption{Token generation throughput under different GPU memory budgets on Mixtral-8x7B. The x-axis shows the memory available for expert caching as a fraction of total expert size.}
    \label{fig:throughput_memory}
\end{figure}

We evaluate the throughput of CRISP under varying GPU memory budgets. Figure~\ref{fig:throughput_memory} shows the token generation throughput as a function of available GPU memory for caching experts.
Under severely constrained memory (10\% of total expert size), CRISP achieves \textbf{8.3} tokens/second, representing a \textbf{3.9}$\times$ speedup over FO (\textbf{2.1} tokens/second) and \textbf{2.1}$\times$ over MO (\textbf{3.9} tokens/second). This improvement stems from two factors: reduced I/O volume due to differentiated pruning, and improved I/O-computation overlap from our pruning-aware scheduler. As memory budget increases, the performance gap narrows but remains significant. At 30\% memory budget, CRISP achieves \textbf{14.7} tokens/second compared to \textbf{8.6} for MO, a \textbf{1.7}$\times$ improvement. Even at 50\% memory budget, CRISP maintains a \textbf{1.4}$\times$ advantage, demonstrating that our approach provides consistent benefits across memory regimes.
Sequence-length throughput results appear in Appendix~\ref{app:sequence_length}.

\subsection{Energy Efficiency Across Edge Platforms}
\label{subsec:energy_efficiency}

\begin{figure}[t]
    \centering
    \includegraphics[width=\linewidth]{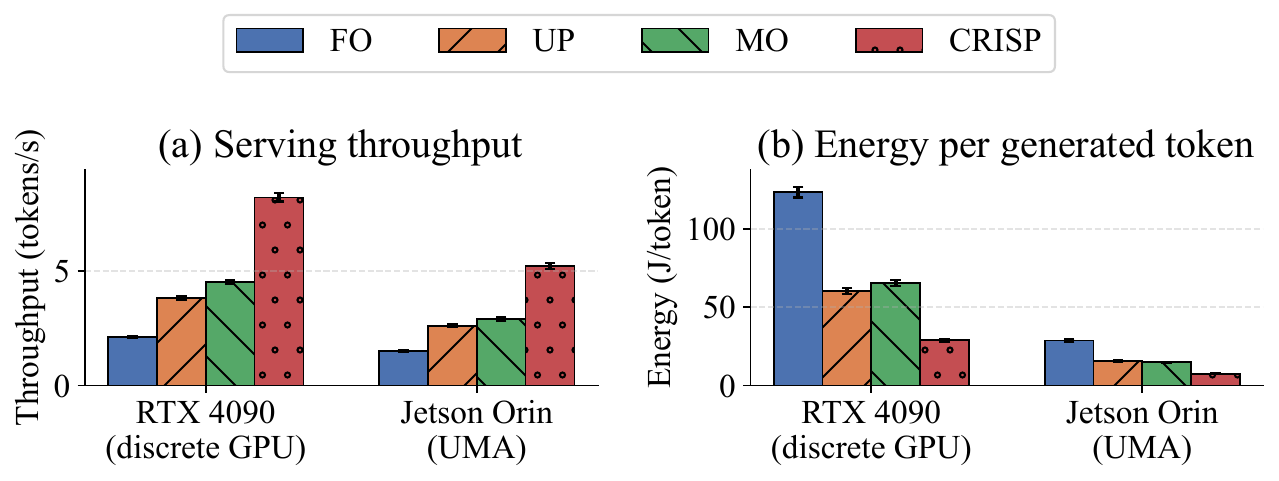}
    \caption{Serving throughput and energy per generated token on discrete-GPU and UMA edge platforms under a 10\% expert-memory budget. Error bars show 95\% confidence intervals.}
    \label{fig:energy_efficiency}
\end{figure}

Figure~\ref{fig:energy_efficiency} compares throughput and energy per generated token on the discrete-GPU and UMA platforms. CRISP reaches 8.2 and 5.2 tokens/s on the RTX 4090 and Jetson Orin, respectively, while consuming 28.9 and 7.5 J/token. Relative to full offloading, this reduces energy per token by 76.7\% on the PCIe system and 73.9\% on the UMA system. On the RTX 4090, the gain comes primarily from reducing and overlapping explicit transfers; on the Jetson Orin, the smaller expert-weight accesses reduce shared-memory bandwidth demand, cache pressure, and memory-fabric contention.
One-time preparation takes 48 minutes, metadata uses less than 2 MB, and scheduling overhead is at most 0.06\% of token latency; see Appendix~\ref{app:system_overhead}.

\subsection{Ablation Study}
\label{subsec:ablation}

We conduct ablation studies to understand the contribution of each component in CRISP.

\subsubsection{Effect of Neuron Reordering.}
\begin{figure}[t]
\centering
\includegraphics[width=.95\columnwidth]{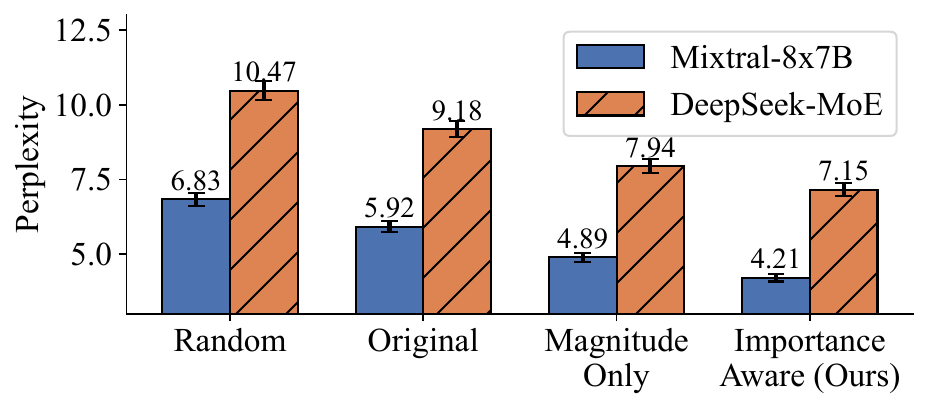}
\caption{Perplexity comparison of different neuron ordering strategies on WikiText-103 at 50\% pruning ratio.}
\label{fig:ablation_reorder}
\end{figure}

We compare our importance-aware reordering against alternative neuron selection strategies: (1) \textit{Random ordering}, which shuffles neurons randomly before truncation; (2) \textit{Original ordering}, which truncates neurons in their original order without reordering; and (3) \textit{Magnitude-only ordering}, which sorts by weight magnitude rather than activation magnitude.
Figure~\ref{fig:ablation_reorder} shows that importance-aware reordering achieves \textbf{4.21} perplexity, significantly outperforming random ordering (\textbf{6.83}) and original ordering (\textbf{5.92}). This confirms that neuron importance varies significantly and that activation-based ranking effectively identifies the most critical neurons to retain.

\subsubsection{Effect of Tiered Execution Policy.}
We evaluate the impact of our hotness-based tiered policy by comparing against uniform execution strategies that apply the same configuration to all combinations.
\begin{figure}[t]
\centering
\includegraphics[width=.95\columnwidth]{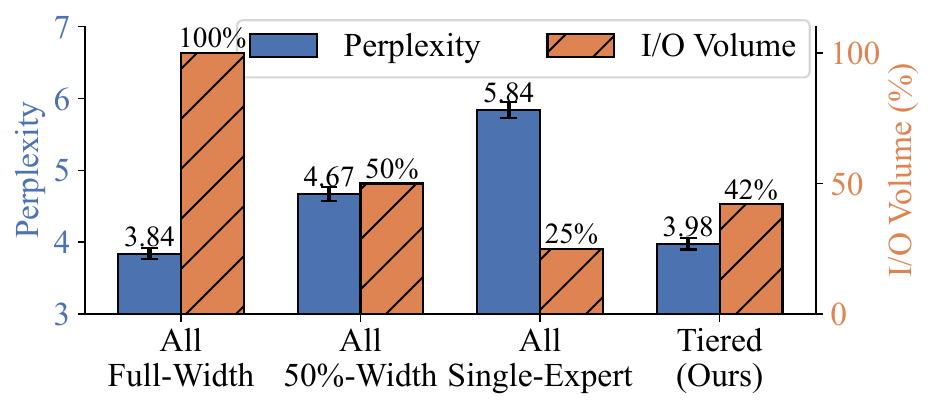}
\caption{Comparison of execution policies on Mixtral-8x7B. PPL denotes WikiText-103 perplexity and Throughput is measured in tokens/second.}
\label{fig:ablation_tier}
\end{figure}
Figure~\ref{fig:ablation_tier} shows that our tiered policy achieves a superior accuracy-efficiency trade-off. Compared to all full-width execution, CRISP reduces I/O volume by \textbf{58}\% while increasing perplexity by only \textbf{0.14}. Compared to uniform 50\%-width pruning, CRISP achieves \textbf{0.69} lower perplexity at similar I/O volume.
At a 75\% I/O budget, asymmetric execution achieves 4.02 perplexity versus 4.18 for symmetric execution; see Appendix~\ref{app:asymmetric_execution}.

\subsubsection{Effect of Pruning-Aware Scheduling.}
\begin{figure}[t]
\centering
\includegraphics[width=.95\columnwidth]{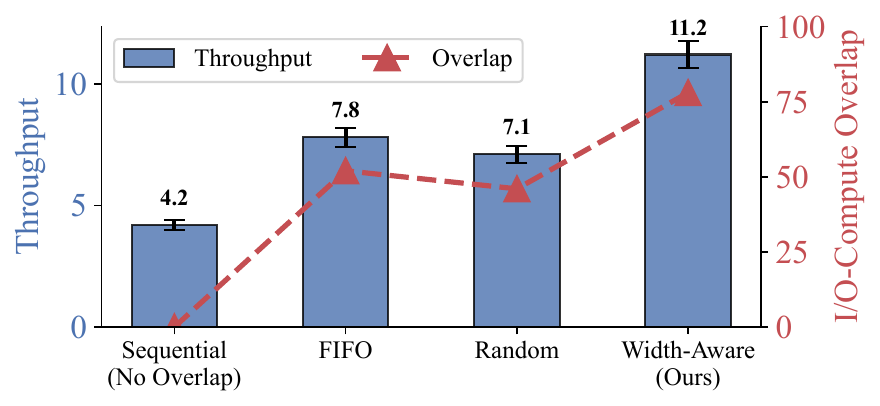}
\caption{Throughput comparison of scheduling strategies on Mixtral-8x7B with 20\% memory budget.}
\label{fig:ablation_scheduling}
\end{figure}

Finally, we evaluate the benefit of our width-aware scheduling strategy by comparing against alternative scheduling policies.
Figure~\ref{fig:ablation_scheduling} shows that width-aware scheduling achieves \textbf{78}\% I/O-computation overlap, compared to \textbf{52}\% for FIFO scheduling. This translates to a \textbf{1.44}$\times$ throughput improvement over FIFO. The key insight is that prioritizing wider experts creates longer computation windows that mask the I/O latency of subsequent narrow experts.
\section{Related Work}

\noindent\textbf{Efficient MoE Inference Systems.}
Prior work has improved MoE inference through system-level optimizations such as expert parallelism, dispatch optimization, and activation-aware caching. PowerInfer~\cite{powerinfer} exploits activation locality to accelerate LLM inference on consumer GPUs by preloading frequently activated neurons. DeepSpeed-MoE~\cite{deepspeed_moe} provides efficient expert parallelism and communication support for large-scale MoE training and inference. Tutel~\cite{tutel} optimizes expert dispatch and execution through a highly tuned system stack. These systems improve expert management and runtime efficiency, but they do not reduce the amount of expert weight data that must be loaded. In contrast, CRISP targets this bottleneck directly through combination-aware pruning.


\noindent\textbf{Neural Network Pruning.}
Pruning and structured submodel selection have been widely explored to reduce computation and adapt model capacity to resource constraints~\cite{pruning_li,pruning_he,chen2025adaptive,tao2026autosub}. For transformers, magnitude-based and gradient-based methods estimate parameter importance and prune accordingly~\cite{pruning_magnitudes,pruning_movement}. Elastic on-device serving also uses neuron reordering to support submodel switching under different latency targets~\cite{yin2025elastic}. Efficient MoE inference additionally requires accounting for sparse, input-dependent routing: expert usage is highly skewed, and an expert's contribution can vary across the combinations in which it appears. CRISP addresses these properties by selecting expert execution widths according to combination frequency and expert role, and coordinating the resulting heterogeneous executions with runtime scheduling.


\noindent\textbf{LLM Quantization.}
Quantization is a widely used approach for reducing LLM memory footprint. Weight-only methods such as GPTQ~\cite{gptq} and AWQ~\cite{awq} compress model weights while preserving higher-precision activations. Other methods, including SmoothQuant~\cite{smoothquant} and Atom~\cite{atom}, quantize both weights and activations to enable more efficient integer computation. Recent MoE serving systems such as D2MoE~\cite{wang2025d2moe} further reduce expert-weight representation cost through quantization. These methods reduce representation cost, but they do not change how much model capacity is retained for different expert combinations. Our approach is orthogonal. CRISP prunes experts based on their usage patterns and can be combined with quantization for further compression. In particular, quantization methods reduces the number of bits used to represent expert weights, whereas CRISP reduces the combination-specific expert capacity that must be accessed and co-designs the resulting heterogeneous widths with runtime scheduling. The two approaches operate along different dimensions and can therefore be combined.


\noindent\textbf{Scheduling and I/O-Computation Overlap.}
At the network level, task scheduling and offloading jointly determine where work is executed and how communication affects completion time~\cite{zhou2024optimizing}. Within a device, overlapping data movement with computation is a common technique for improving system throughput. GPipe~\cite{gpipe} uses micro-batching to pipeline computation across stages, while FlashAttention~\cite{flashattention} improves efficiency through kernel-level optimization and memory-aware execution. For LLM inference, speculative decoding~\cite{speculative_decode} and continuous batching~\cite{orca} improve resource utilization and throughput. CRISP differs from these approaches in both source and use of heterogeneity. Rather than relying on pipeline structure or request-level batching, CRISP creates heterogeneous expert execution times through differentiated pruning and exploits that heterogeneity with a width-aware scheduler to overlap expert loading with computation.

\vspace{-.1in}
\section{Conclusion}

This paper shows that the main challenge in edge MoE serving is not only
reducing expert-weight access, but deciding where expert capacity can be reduced in a way that matches how experts are actually used at runtime.
Our results suggest that MoE pruning should be guided by expert combinations rather than by experts in isolation, and that the execution heterogeneity created by differentiated pruning should be treated as a systems opportunity rather than a side effect.
CRISP realizes this perspective through a co-design in which importance-aware reordering enables multiple execution widths, combination-aware pruning selects among them based on combination frequency and expert role, and pruning-aware scheduling exploits the resulting heterogeneity to reduce exposed I/O latency.
This co-design substantially improves throughput under tight memory budgets while maintaining a favorable quality-latency trade-off.
More broadly, our study highlights that efficient deployment of sparse models on edge platforms requires jointly reasoning about model structure, access patterns, and execution behavior.


\begin{acks}
This work was supported in part by the Major Basic Research Program of
Shandong Provincial Natural Science Foundation under Grant ZR2025ZD18, the
National Natural Science Foundation of China under Grant 62572280, U24A20244, City University of Hong Kong 9229212, and Hong Kong RGC C1043-24GF.
\end{acks}

\bibliographystyle{ACM-Reference-Format}
\bibliography{sample-base}

\clearpage
\appendix
\section{Additional Evaluation Results}

\subsection{Tier-Conditioned Accuracy}
\label{app:tier_conditioned_accuracy}

\begin{figure}[H]
    \centering
    \includegraphics[width=\linewidth]{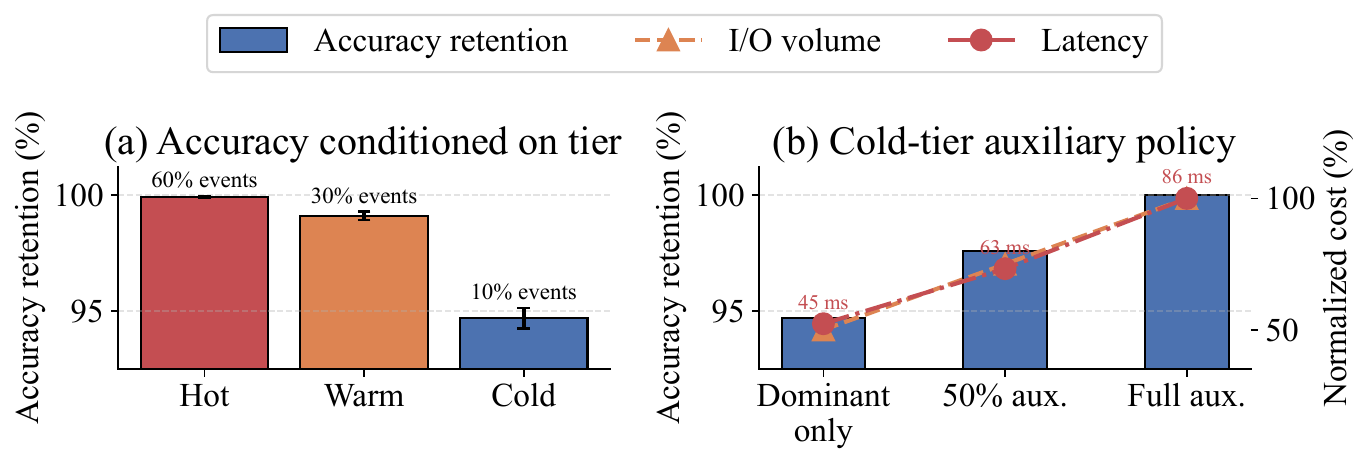}
    \caption{Tier-conditioned accuracy and Cold-tier auxiliary-expert analysis. The left panel reports accuracy retention within each execution tier and its share of routing events. The right panel compares accuracy retention, I/O volume, and latency when the Cold tier executes only the dominant expert, retains 50\% of the auxiliary expert, or retains the full auxiliary expert.}
    \label{fig:tier_conditioned_accuracy}
\end{figure}

Dataset-level averages can obscure whether aggressive execution disproportionately affects a particular tier. We therefore report accuracy retention conditioned on Hot, Warm, and Cold routing events and further vary the amount of auxiliary-expert capacity retained for Cold combinations. Figure~\ref{fig:tier_conditioned_accuracy} summarizes both analyses.

Hot and Warm combinations retain 99.9\% and 99.1\% of full-width accuracy, respectively, while Cold combinations retain 94.7\%. Because Cold combinations account for only 10\% of routing events, their larger conditional degradation has a limited effect on dataset-level accuracy. Within the Cold tier, executing only the dominant expert reduces I/O volume by 50\% and latency from 86 ms to 45 ms, at the cost of reducing accuracy retention from 100\% to 94.7\%; retaining 50\% of the auxiliary expert provides an intermediate operating point. This result makes the effect of skipping the Cold-tier auxiliary expert explicit and allows the policy to be selected according to the desired quality-latency trade-off.

\subsection{Robustness to Workload Shift}
\label{app:workload_shift}

\begin{figure}[H]
    \centering
    \includegraphics[width=\linewidth]{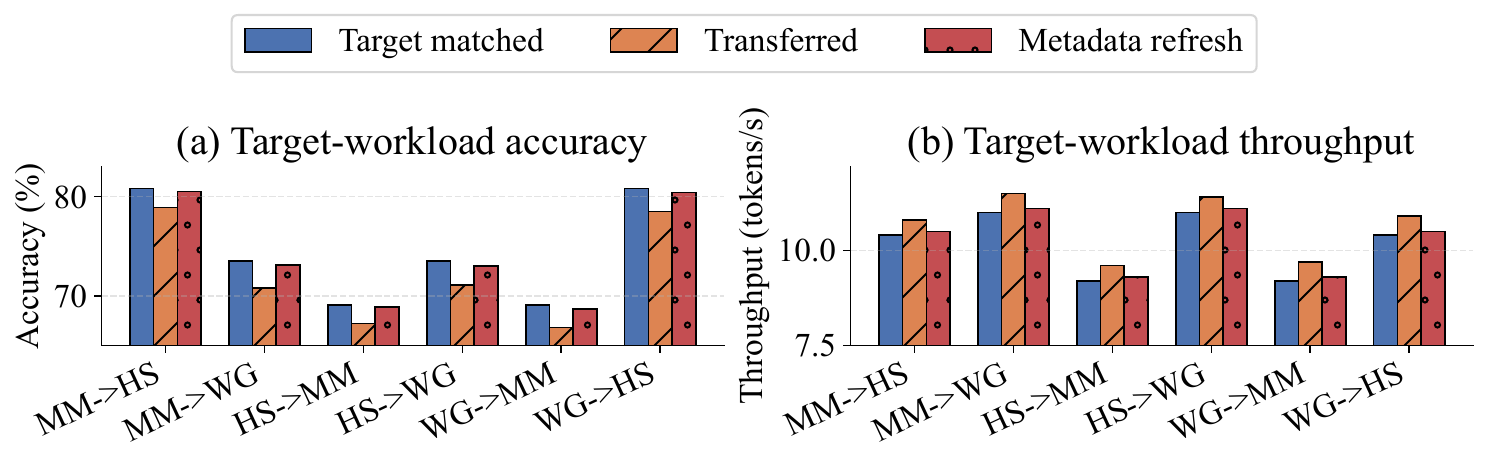}
    \caption{Cross-domain robustness of CRISP when profiling metadata is transferred between MMLU (MM), HellaSwag (HS), and WinoGrande (WG). Target matched uses metadata profiled on the evaluation workload, Transferred directly applies source-workload metadata, and Metadata refresh updates the metadata on the target workload.}
    \label{fig:cross_domain_shift}
\end{figure}

We evaluate the robustness of CRISP when profiling metadata collected on one workload is applied to another. Figure~\ref{fig:cross_domain_shift} reports all cross-domain transfers among MMLU (MM), HellaSwag (HS), and WinoGrande (WG), comparing metadata profiled on the target workload (Target matched), metadata directly transferred from the source workload (Transferred), and metadata updated using the target workload (Metadata refresh).

Directly transferring metadata limits the accuracy decrease to at most 2.7 percentage points across the evaluated workload pairs, while maintaining comparable or slightly higher throughput. Refreshing the profiling metadata recovers accuracy to within 0.5 percentage points of the target-matched setting and retains similar throughput. These results show that CRISP degrades gracefully under workload shift and can recover most of the quality loss by updating its profiling metadata for the new workload.

\subsection{Throughput Across Sequence Lengths}
\label{app:sequence_length}

\begin{figure}[H]
    \centering
    \includegraphics[width=.9\columnwidth]{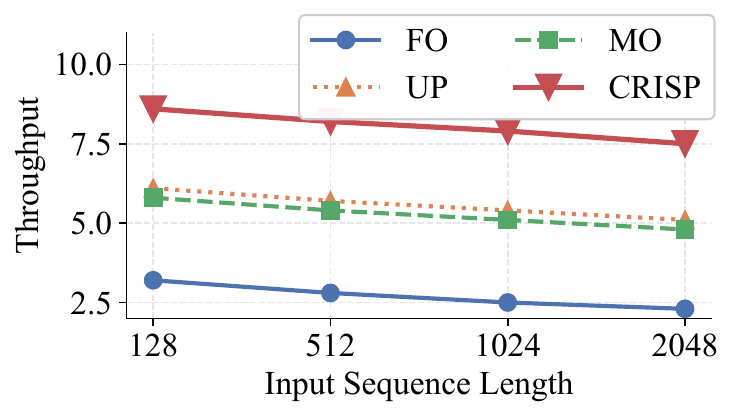}
    \caption{Throughput (tokens/second) across different sequence lengths on Mixtral-8x7B with 20\% memory budget.}
    \label{fig:throughput_breakdown}
\end{figure}

We evaluate throughput across input sequence lengths from 128 to 2,048 tokens on Mixtral-8x7B under a 20\% memory budget. Figure~\ref{fig:throughput_breakdown} shows that CRISP maintains its advantage across all evaluated lengths, with greater improvements for longer sequences as the cumulative benefit of reduced I/O becomes more pronounced.

\subsection{System Overhead Analysis}
\label{app:system_overhead}

We analyze the overhead of CRISP, including both offline preparation costs and runtime overhead.

\subsubsection{Offline Preparation Overhead.}
The importance-aware neuron reordering and combination classification are performed offline and incur no runtime cost. Table~\ref{tab:offline_overhead} reports the offline preparation time for these components.
The Importance Computation (IC) requires a forward pass over the calibration dataset (\textbf{1024} samples) while recording intermediate activations, completing in \textbf{18} minutes for Mixtral-8x7B. The Weight Reordering (WR) involves permuting the weight matrices according to the computed importance ranking, which takes \textbf{7} minutes. Combination Profiling (CP) requires running inference on the calibration dataset while recording activated combinations, which takes \textbf{15} minutes for Mixtral-8x7B. The Dominant Expert Identification (DEI) adds \textbf{8} minutes for ablation-based analysis. The total offline time of \textbf{48} minutes is a one-time cost amortized over all subsequent inferences.
The storage overhead for reordered weights is negligible: the reordered model occupies the same disk space as the original, as we only permute existing values without adding new parameters. The only additional storage is the tier assignment table and dominant expert mapping, which together consume less than \textbf{2} MB. For Mixtral-8x7B, \textbf{14.3}\% of combinations are classified as Hot, \textbf{32.1}\% as Warm, and \textbf{53.6}\% as Cold. However, due to the power-law distribution, Hot combinations account for \textbf{61.8}\% of actual activations at runtime.

\begin{table}[H]
    \centering
    \caption{Offline preparation overhead for CRISP. IC: Importance Computation, WR: Weight Reordering, CP: Combination Profiling, DEI: Dominant Expert Identification.}
    \label{tab:offline_overhead}
    \begin{tabular}{lcc}
        \toprule
        \textbf{Component} & \textbf{Mixtral-8x7B} & \textbf{DeepSeek-MoE} \\
        \midrule
        IC & \textbf{18} min & \textbf{12} min \\
        WR & \textbf{7} min & \textbf{4} min \\
        CP & \textbf{15} min & \textbf{22} min \\
        DEI & \textbf{8} min & \textbf{10} min \\
        \midrule
        Total Offline Time & \textbf{48} min & \textbf{48} min \\
        \bottomrule
    \end{tabular}
\end{table}

\subsubsection{Runtime Scheduling Overhead.}
The pruning-aware scheduler introduces runtime overhead for maintaining priority queues and making scheduling decisions. We measure this overhead by comparing against a baseline scheduler without width-awareness. As shown in Table~\ref{tab:scheduling_overhead}, the total scheduling overhead is \textbf{48.1} $\mu$s per token for Mixtral-8x7B, representing only \textbf{0.05}\% of the total token generation latency. The Tier Lookup (TL) is implemented as a GPU-resident table access and completes in \textbf{0.8} $\mu$s. Stream Coordination (SC) adds \textbf{45} $\mu$s but enables I/O prefetching that saves significantly more time. Overall, the scheduling overhead is negligible compared to the performance gains from improved I/O-computation overlap.

\begin{table}[H]
    \centering
    \caption{Runtime scheduling overhead per token. TL: Tier Lookup, PQO: Priority Queue Operations, SC: Stream Coordination.}
    \label{tab:scheduling_overhead}
    \begin{tabular}{lcc}
        \toprule
        \textbf{Component} & \textbf{Mixtral-8x7B} & \textbf{DeepSeek-MoE} \\
        \midrule
        TL & \textbf{0.8} $\mu$s & \textbf{1.2} $\mu$s \\
        PQO & \textbf{2.3} $\mu$s & \textbf{4.7} $\mu$s \\
        SC & \textbf{45} $\mu$s & \textbf{38} $\mu$s \\
        \midrule
        Total overhead & \textbf{48.1} $\mu$s & \textbf{43.9} $\mu$s \\
        Token latency & \textbf{0.05}\% & \textbf{0.06}\% \\
        \bottomrule
    \end{tabular}
\end{table}

\subsection{Asymmetric Execution}
\label{app:asymmetric_execution}

For Warm-tier combinations, we compare our asymmetric execution (full-width dominant + narrow auxiliary) against symmetric alternatives.

\begin{table}[H]
    \centering
    \caption{Comparison of Warm-tier execution strategies on Mixtral-8x7B.}
    \label{tab:ablation_asymmetric}
    \begin{tabular}{lcc}
        \toprule
        \textbf{Warm-Tier Strategy} & \textbf{PPL$\downarrow$} & \textbf{I/O Volume} \\
        \midrule
        Both Full-Width & 3.84 & 100\% \\
        Both 75\%-Width & \textbf{4.18} & 75\% \\
        Both 50\%-Width & \textbf{4.67} & 50\% \\
        Asymmetric (100\% + 50\%) & \textbf{4.02} & 75\% \\
        \bottomrule
    \end{tabular}
\end{table}

At the same 75\% I/O budget, asymmetric execution achieves \textbf{4.02} perplexity compared to \textbf{4.18} for symmetric 75\%-width execution. This validates our observation that experts play asymmetric roles within a combination, and that preserving full capacity for the dominant expert is more important than balanced pruning.

\end{document}